# Analytical and Bootstrap Confidence Intervals for Double Machine Learning: Simulation Studies and an Application to Rural-Urban Differences in Obesity Prevalence

Haozheng Xu[1,a], Siyuan Ma[1,b], and Qingyan Xiang[1,c,*]

[1]*Department of Biostatistics, Vanderbilt University Medical Center,*
[a]*haozheng.xu@vanderbilt.edu*, [b]*siyuan.ma@vumc.org*, [c]*qingyan.xiang@vumc.org*
**: Corresponding author*

**Abstract**

Double Machine Learning (DML) is a popular treatment effect estimation framework that allows flexible machine learning methods to be used for nuisance parameter estimation. In practice, applied researchers must choose among many learners, and how this choice affects the variance estimation of DML is not well characterized. We conduct a comprehensive simulation study to compare the coverage probability of DML confidence intervals using different learners. In this study, we compare (1) analytical confidence intervals derived by DML theory versus (2) bootstrap confidence intervals. We use a set of learners including ordinary least squares, LASSO, Random Forest, LightGBM, and Neural Networks under different data generation settings. Evaluation is based on bias, confidence interval width, and coverage probability. Our results show substantial variability in coverage performance across analytical and bootstrap confidence intervals, highlighting that learner choice plays a critical role in reliable DML inference. Surprisingly, we find in many settings, as sample size increases, the coverage probability of both DML analytical and bootstrap confidence intervals decrease. We further examine coverage in a county-level analysis of rural–urban differences in obesity prevalence, where performance again varies by learner, and a higher level of rurality has a positive effect on obesity prevalence.

## 1. Introduction

Double (debiased) Machine Learning (DML) is an estimation method that has gained popularity in multiple areas of research[1,2]. Initially introduced by Chernozhukov[3], the DML method can estimate structural parameters with the Partially Linear Regression (PLR) model. Because DML incorporates the use of flexible machine learning algorithms, it can relax modeling specification requirements against traditional parametric approaches, especially for complicated settings[4]. Given the strong potential of application, it is crucial to understand its behavior under various data generation settings, performance among different learners, and how it performs in practical analysis.

In recent years, many scholars have applied and evaluated DML in empirical studies and simulation settings. Yang et al.[5] apply DML using gradient boosting to estimate causal effects in high-dimensional observational data, showing that DML produces more stable estimates than traditional regression methods. Fuhr[6] conducts a review of multiple machine learning methods to simulate model performance under different data generating mechanisms. Others focus on the DML estimator's finite-sample properties: Saco shows that the bias of the DML estimator can be magnified due to certain conditions[7]; Ballinari[8] shows there is marked difference in estimated RMSE and bias while using different learners; and Bach[9] shows that selection and tuning of nuisance learners can substantially affect estimator bias and RMSE. Additionally, DML is shown to effectively estimate binary[3] and continuous treatment[10]; and it has been extended to estimate heterogeneous treatment effects[11,12]. All these studies show its flexibility to adapt to different machine learning methods and people's interest in its behaviors.

While the existing literature effectively shows the strength of DML and evaluates its behavior, relatively less attention is paid to its inference properties, especially the coverage probability of its analytical confidence interval. In practice, many empirical studies rely on standard errors based on the DML asymptotic theory. However, we have little knowledge of how coverage probabilities differ across analytical variance estimates and bootstrap variance estimates across different data settings and learners[13]. Therefore, in this study, our goal is to evaluate the difference between analytical and bootstrap variances and their corresponding confidence interval coverages. Comparisons are made across multiple machine learning learners and data-generating processes. We hope this study can help the DML framework to be better understood, used, and interpreted for data analyzing and practical statistical inference.

## 2. Methods

### 2.1 Overview

This study focuses on evaluating the coverage probability of both analytical and bootstrap variance estimators of the DML partially linear model, and **Figure 1** illustrates the overall workflow. We begin by simulating different Data Generating Processes (DGP). Then, using the DML framework, we estimate the treatment and outcome models with a range of machine learning algorithms in each setting. We subsequently construct confidence intervals for the treatment effect using two inference strategies: (1) analytical variance estimation derived from the asymptotic theory and (2) bootstrap-based inference. Performance of these two variance estimation is assessed through simulation metrics such as analytical bias, coverage probability, analytical and empirical standard errors, and confidence interval width. Finally, we apply the DML partially linear model to a real dataset to evaluate practical performance. All the comparisons will be permuted from different DGPs, machine learning methods, and CI computing methods, to ensure a comprehensive review of our cases.

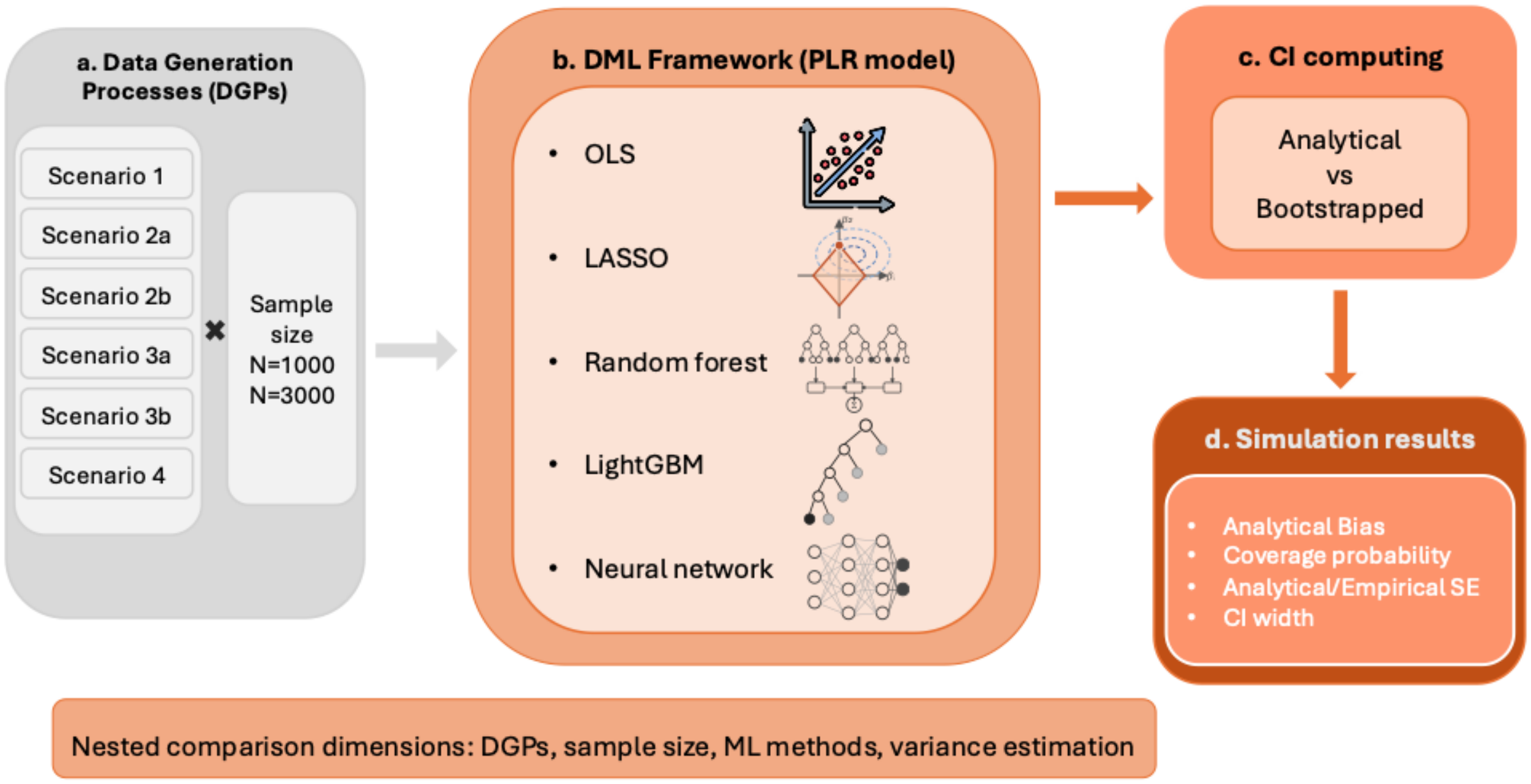

**Figure 1**: Overall workflow of this work: Abbreviations: DML = Double Machine Learning; PLR = Partially Linear Regression; OLS = Ordinary Least Squares; LASSO = Least Absolute Shrinkage and Selection Operator; LightGBM = Light Gradient Boosting Machine; CI = Confidence Interval; SE = Standard Error; DGP = Data Generating Process.

## 2.2 Review on Double (Debiased) Machine Learning

### 2.2.1 Partially Linear Regression (PLR)

In this study, we focus on the DML partially linear regression (PLR) model[3], and we briefly review this method. Define $D$ as the treatment or exposure of interest and $\theta$ as its effect on the outcome $Y$. $X$ is a set of confounders that influence both the treatment and the outcome. The PLR model takes the form

$$Y = D\theta + g(X) + \varepsilon, E(\varepsilon|X) = 0,$$

$$D = m(X) + \zeta, E(\zeta|X) = 0,$$

where $g(X)$ is the nuisance function from the confounding variables to the outcome; $m(X)$ is the function between confounders and treatment; $\varepsilon$ and $\zeta$ are random error terms. Without correctly accounting for the confounders, simply using machine learning for estimating on the treatment effect of $D$ on $Y$ can cause overfitting and biased estimation.

Here, the PLR model uses a "partialling out" approach that comes from a double residual method[14]. Formally, it is estimating $\hat{\theta}$ by:

1. Modeling $Y_i$ on $X_i$ to get residuals $\tilde{Y}_i = Y_i - \hat{g}(X_i)$.
2. Modeling $D_i$ on $X_i$ to get residuals $\tilde{D}_i = D_i - \hat{m}(X_i)$.
3. Regressing $\tilde{Y}_i$ on $\tilde{D}_i$ to get the estimated treatment effect.

This regression-on-residuals formulation provides the same estimate as the full regression and guarantees that the parameter $\theta$ can be estimated even if we have little knowledge of the functional structures.

However, the issue with this mechanism is that if the data is not simple, the estimation fails easily with model misspecification or overfitting. Specifically, when the functions $g(X_i)$ and $m(X_i)$ have complex structures, or if the machine learning method needs tuning, much more effort is needed to specify the model structure correctly and produce unbiased estimation.

To overcome the issue, Double Machine Learning algorithms are used to estimate $g(X_i)$ and $m(X_i)$. The PLR framework further uses a cross-fitting mechanism to prevent overfitting. We first split the dataset into $K$ folds, specifying one of the partitioned folds as our testing dataset, and train the nuisance model on the remaining training sets. The trained model will then be used to generate estimated $\hat{g}(X)$ and $\hat{m}(X)$, with which we residualize the observations from the testing dataset. This procedure ensures that each observation and residualized component is predicted using models that were not trained on that observation, therefore we reduce overfitting bias. It is also advised to repeat the whole process $M$ times (each time with different partitions) to further ensure robustness[15 3].

### 2.2.2 Analytical Variance Estimation

We used analytical confidence intervals based on the asymptotic theory of DML[3]. Under standard regularity conditions, the DML estimator satisfies:

$$\sqrt{(N)}\left(\hat{\theta} - \theta_0\right) \rightarrow N(0, \sigma^2).$$

The key idea is to construct a score function $\psi$ that is insensitive (orthogonal) to small errors in estimating the nuisance parameters $g(X_i)$ and $m(X_i)$. For the partially linear DML model, the score function can be written as

$$\psi(W_i; \hat{\theta}, \hat{\eta}) = \left(D_i - \hat{m}(X_i)\right)\left[Y_i - \hat{g}(X_i) - \hat{\theta}\left(D_i - \hat{m}(X_i)\right)\right],$$

where $W_{\mathrm{i}} = (Y_{\mathrm{i}}, D_{\mathrm{i}}, X_{\mathrm{i}})$ and $\hat{\eta} = (\hat{g}, \hat{m})$ are the nuisance parameters estimated using machine learning methods.

The variance of $\hat{\theta}$ can be estimated in accordance with the asymptotic theory of DML. Empirically, after we estimated the score function $\hat{\psi}$, when $\theta$ is a scalar, the variance formula is

$$\hat{\sigma}^2 = \hat{J_0}^{-2} \frac{1}{N} \sum_{i=1}^{N} \hat{\psi}_i^2,$$

where $\hat{J}_0$ is estimated by

$$\hat{J}_0 = \frac{1}{N} \sum_{i=1}^{N} \left(D_i - \hat{m}(X_i)\right)^2.$$

This leads to the estimated standard error $\widehat{se}(\hat{\theta}) = \sqrt{(\hat{\sigma}^2/N)}$ and the corresponding

$(1-\alpha) \times 100\%$ confidence interval $\hat{\theta} \pm z_{1-\alpha/2}\, \widehat{se}(\hat{\theta})$.

In our study, we compute the DML analytical confidence interval using Bach and Chernozhukov's DoubleML[9] package. We implement the above variance estimation procedures with $K = 5$ cross-fitting folds repeated 10 times, with the median taken across repetitions to reduce sensitivity.

### 2.2.3 Bootstrap Variance Estimation

In addition to the analytical variance, we also evaluate the performance of DML with bootstrapping[16]. This allows us to construct quantile-based confidence intervals based on the distribution of DML analytical fits, without relying only on asymptotic approximations or analytical standard errors. Formally, we draw $B$ bootstrap resamples of size $N$ with replacement, and within each bootstrapped sample, compute the estimated $\hat{\theta}^{(i,b)}$ using DML. Next, we calculate the mean of the bootstrap estimates $\bar{\theta}^{(i)} = \frac{1}{B}\sum_{b=1}^{B} \hat{\theta}^{(i,b)}$ and observe the overall distribution of estimates, and eventually we get the percentile confidence interval $\left[0.025\text{th percentile of } \hat{\theta}^{(i,b)}, 0.975\text{th percentile of } \hat{\theta}^{(i,b)}\right]$ from the bootstrap estimates. In all simulations, we set $B = 1000$.

## 2.3 Machine Learning Algorithms

In this study, we implement the DML framework with some commonly used algorithms including ordinary least squares (OLS), LASSO[17], Random Forest[18], LightGBM[19], and single-layer Neural Networks. These methods were chosen to represent different modeling strategies, from linear and sparse regression to nonlinear tree-based models and Neural Networks. While hyperparameter tuning is incorporated for selected methods, others are run with default settings from the package to reflect common usage. The reason is that our primary goal is not to identify optimal hyperparameter sets under different settings, but rather to compare the performance of DML and bootstrap coverage probability in different data-generating processes and algorithms.

The learners considered in this study typically require hyperparameter tuning. To ensure fair comparison across methods and an accurate use of the DML framework, hyperparameter tuning was conducted using grid searching within each data scenario. In this study, we perform tuning for Neural Networks and LightGBM, because of their unstable performance in multiple cases. More details of tuning parameters and tuning results are shown in Appendix A.

### 2.4 Software Implementation

For this study, we used the statistical software R to do all the computations. The use of Double Machine Learning comes from the R package DoubleML[15]. All simulation studies, including data generation, model fitting, and inference evaluation, were performed in R with specified seeds to ensure reproducibility. All machine learning algorithms are carried out using the mlr3[20] package in R to ensure comparability. Results were stored and processed using standard R data structures (rds files), and visualization was conducted using packages such as ggplot2[21]. The full workflow was integrated into a reproducible research pipeline using Quarto, enabling seamless generation of tables, figures, and manuscript content.

## 3. Simulation Settings

### 3.1 Data Generating Process (DGP)

In the simulation study, we have different DGPs that represent different levels of complexity, and our DGPs align with the PLR framework:

$$Y = D\theta + g(X) + \varepsilon,$$

$$D = m(X) + \zeta.$$

In our simulation, we have six scenarios with four different specifications of $g(X)$ and $m(X)$ to generate $Y$ and $D$. In particular, $D$ is continuous in scenarios 1, 2a, 2b, and 4, and $D$ is binary in scenarios 3a and 3b. Table 1 shows the overall setting of data generating mechanisms. Since scenario 4 includes high-dimensional data and the computation burden is high, we evaluate the model performance based on 300 datasets instead of 500.

Additionally, we also introduce treatment of different strengths to evaluate how estimation changes with different levels of treatment effect. Scenario $2a$ and $2b$ have the exact same settings, except that $2a$ has a treatment effect of 0.6, while $2b$ has a treatment effect of 5.

Similarly, Scenario $3a$ has a treatment effect of 0.6, and Scenario $3b$ has a treatment effect of 5. In the following subsections, we show the detailed specification of the data generating function for $g(X)$ and $m(X)$ in each scenario.

| *Scenario* | *Treatment* | *Covariates* | *Covariate complexity* | *Dataset* | *Sample* | *Parameter* |
|---|---|---|---|---|---|---|
| 1 | Continuous | 6 | Linear | 500 | $N = 1000/3000$ | $\theta = 0.6$ |
| $2a$ | Continuous | 6 | Nonlinear | 500 | $N = 1000/3000$ | $\theta = 0.6$ |
| $2b$ | Continuous | 6 | Nonlinear | 500 | $N = 1000/3000$ | $\theta = 5$ |
| $3a$ | Binary | 10 | Nonlinear | 500 | $N = 1000/3000$ | $\theta = 0.6$ |
| $3b$ | Binary | 10 | Nonlinear | 500 | $N = 1000/3000$ | $\theta = 5$ |
| 4* | Continuous | 10 + 50 (noise) | Quasi–High-D | 300* | $N = 1000/3000$ | $\theta = 5$ |

**Table 1: Overview of Simulation Scenarios**. "Dataset" refers to the number of simulation replications. "Sample" refers to the number of observations per replication. "Parameter" refers to the true treatment effect $\theta$ used in each scenario. * The number of replications of Scenario 4 is reduced to 300 due to computational burden.

### 3.1.1 Scenario 1: Linear, Continuous Treatment

In the first scenario, the nuisance functions take a simple linear case:

$$g_1(X) = 1.2X_1 + 0.5X_2 - 0.5X_3 + X_4 + X_5 - X_6,$$

$$m_1(X) = 0.8X_1 - 0.6X_2 + X_3 + 0.2X_4 - 0.5X_5.$$

This scenario serves as a baseline to compare with other more complicated scenarios. Covariates $X_1, \dots, X_5$ were independently sampled from a standard normal distribution.

### 3.1.2 Scenario 2: Nonlinear, Continuous Treatment

In the second scenario, the nuisance functions follow a form with nonlinear, non-smooth setting with indicator functions, interaction terms, and quadratic terms.

$$g_2(X) = 1.2X_3 + 0.5X_4^2 + 0.5(X_5X_6) + X_2 + X_1 + X_1X_2 + 2\left(I_{X_1>0} + I_{X_1<0}\right) + 2\left(I_{X_6>0} + I_{X_6<0}\right),$$

$$m_2(X) = 0.8sin(X_1) - 0.6log(|X_2| + 1) + I_{X_6>0} + 0.5(X_1X_3) + X_4X_5.$$

This scenario is used to evaluate the robustness of the model under nonlinear and non-smooth cases. Again, Covariates $X_1, \ldots, X_5$ come from a simple multivariate normal distribution, with each component independently sampled from a standard normal distribution.

### 3.1.3 Scenario 3: Nonlinear, Binary Treatment

Scenario 3 has a binary treatment variable that is generated from a Bernoulli distribution.

$$D \sim Bernoulli\big(\pi(X)\big), \pi(X) = logit^{-1}\big(m_3(X)\big),$$

where $\pi(X)$ is the probability of treatment assignment defined by the nonlinear function $m_3(X)$.

$$g_3(X) = -2\big(I(X_1 < 0) + I(X_1 > 0)\big) + \big(-I(X_2 < 1) + I(X_2 > 1)\big) + 2X_3 + 2X_5 + X_6 + X_7 - 2X_9 - 0.5X_{10} + 2X_3X_4 + 2X_5X_{10} + 2X_5^2 + 2X_9^2,$$

$$m_3(X) = 1.3X_1X_2 + 0.7{X_2}^2 - 0.4X_3 + exp(X_4) + 1.5X_7X_9 - 1.5X_{10}.$$

This formulation distinguishes Scenario 3 from Scenarios 1 and 2, where we had continuous treatment. The outcome model is also nonlinear, including indicator functions, quadratic terms, exponentials, and interactions. This scenario is created to evaluate performance under binary treatment and complex data structures. For each simulated dataset, 10 baseline covariates were generated independently. Specifically, $X_1, \ldots, X_5$ were generated as independent standard normal variables, $X_j \sim N(0,1)$ for $j = 1, \ldots, 5$, while $X_6, \ldots, X_{10}$ were generated as independent Bernoulli variables with success probabilities $0.1, 0.3, 0.5, 0.7$, and $0.9$, respectively.

### 3.1.4 Scenario 4: Quasi–High-Dimensional, Continuous Treatment

Scenario 4 extends complexity by introducing a Quasi–High-Dimensional setting with 60 covariates, of which 10 are informative and 50 are noise. The treatment is continuous and is generated from a complex setting. The outcome model follows the same nonlinear structure as Scenario 3, to preserve some similarity while adding the impact of dimensionality.

$$g_4(X) = g_3(X),$$

$$m_4(X) = 2X_1I(X_1 > 0) - X_1I(X_1 < 0) + sin(X_2) + {X_3}^2I(X_3 > 0.5) - 2X_3I(X_3 < 0) + 0.5X_4X_5 + \log(|X_6| + 1).$$

For each simulated dataset, the covariate vector $X = (X_1, \ldots, X_p)$ with $p = 60$ was generated from a multivariate normal distribution with mean zero and a structured covariance matrix. Specifically,

$$X \sim N_p(0, \Sigma),$$

where the diagonal entries of $\Sigma$ are 1, and the first 10 signal covariates are correlated with the remaining 50 noise covariates through a constant correlation parameter $\rho = 0.3$. Since the block-correlation structure does not guarantee positive definiteness for all values of $\rho$, the covariance matrix $\Sigma$ was adjusted to the nearest positive definite matrix using the method of Higham[22], implemented via the nearPD() function in R. This ensures valid draws from the multivariate normal distribution while preserving the intended correlation structure as closely as possible.

### 3.2 Evaluation

Our primary objective is to compare inference obtained from DML using analytical asymptotic confidence intervals with inference obtained from DML bootstrap confidence intervals. For each simulated dataset ($\mathrm{N} = 1000/3000$), we use the DoubleML[15] package to implement the DML framework with 5-fold cross-fitting[3], repeated 10 times to increase the robustness of estimation[4]. Nuisance functions ($g(X)$ and $m(X)$) were estimated using OLS, LASSO, Random Forest, LightGBM, and single-layer Neural Networks. A total of 500 independent replications were conducted for each of the six simulation scenarios, except for scenario 4, for which 300 replications were used.

Performance was evaluated along four dimensions. First, point estimation accuracy is assessed through empirical bias, defined relative to the true treatment effect $\theta$. Second, variation is measured by comparing the empirical standard deviation across replications with the average analytical standard errors, along with the confidence interval width to show the precision of estimation. Third, we calculate the coverage probability of nominal 95% confidence intervals in both the DML analytical and bootstrapping methods. Together, these criteria provide a comprehensive comparison of DML and bootstrap inference across different levels of DGP and model complexity.

## 4. Simulation results

### 4.1 Overall review

This section shows the simulation results from different settings. Figure 2 presents boxplots of estimated treatment effects across six scenarios for sample size $N = 1000$ and 3000. Each box summarizes the distribution of estimates obtained from 500 replications (with the exception of scenario 4, 300 instead), the dashed horizontal line indicate the true treatment effect ($\theta = 0.6$ or 5, depending on the scenario). Red boxes are the analytical estimation, and blue boxes correspond to the bootstrap estimates. Across most settings, analytical and bootstrap estimates share similar centers and spreads, but there are some nuances across cases and methods. Figure 3 displays the coverage probability in the form of a bar chart. It shows a lack of definitive difference between analytical and bootstrap coverage probability. To summarize key takeaways from Figure 2 and Figure 3:

1. There can be very different estimation performance within each scenario comparing

different methods, and across scenarios when comparing one single method.

2. When the scenarios are linear and simple, OLS and LASSO tend to perform well. This shows that complicated models are not always the best to the analysis; a suitable model matters more, even if the model is the simplest OLS model.
3. Correct model specification matters a lot, as we observe from the varying performances of OLS and LASSO while used at scenarios of different complexity.
4. When it goes to the nonlinear cases, Machine Learning models would do well in weak signal cases, but when the signal strength is strong, it can be hard for Machine Learning models to estimate properly.
5. When the sample size increases, the coverage probability of DML methods does not necessarily improve. This happens as the results show that DML does not necessarily decrease estimation bias, but the variance estimates still shrink, which leads to the decrease of the coverage.

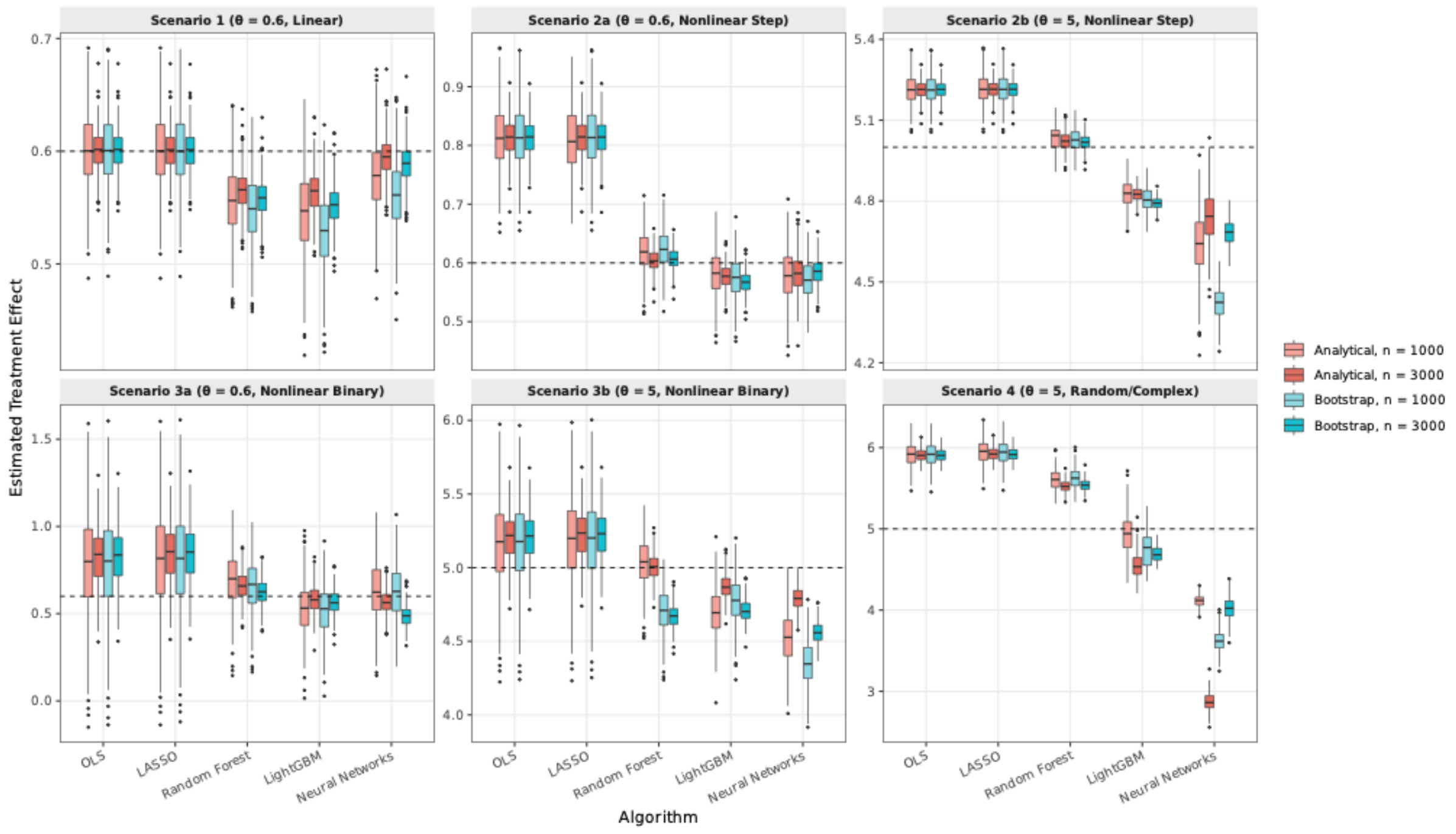

**Figure 2**: Distribution of Estimated Treatment Effects by Scenario and Learner*:* Distribution of estimated treatment effects across six simulation scenarios and two sample sizes ($N$ = 1,000 and $N$ = 3,000). Each panel corresponds to one scenario (title indicates true treatment effect $\theta$ and DGP type). The horizontal axis shows the five machine learning learners described in Section 2.3. The dashed horizontal line marks the true treatment effect $\theta$. Red boxes indicate analytical confidence interval estimates and blue boxes indicate bootstrap estimates; lighter shades correspond to $N$ = 1,000 and darker shades to $N$ = 3,000.

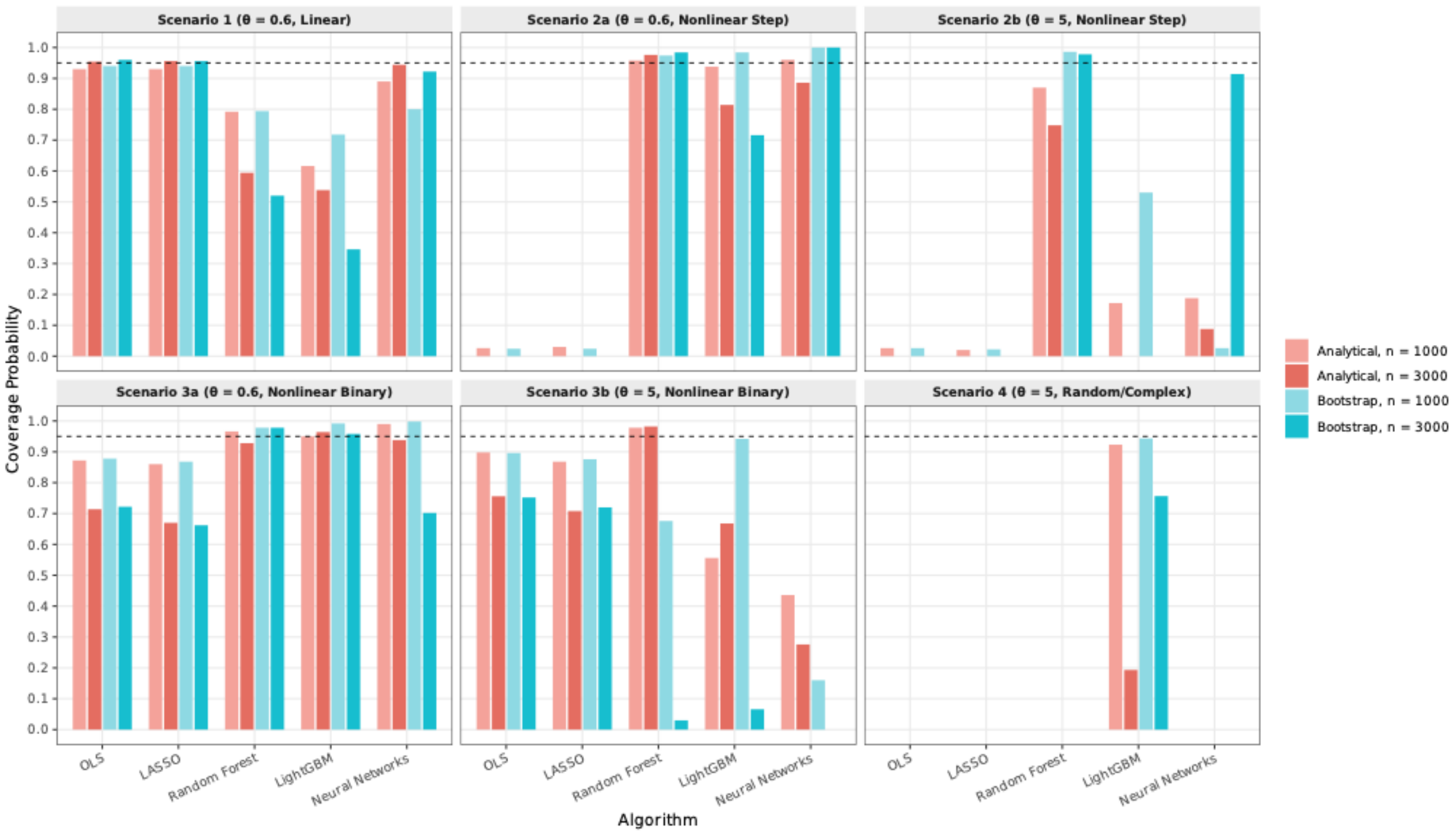

**Figure 3**: Coverage probability of analytical and bootstrap confidence intervals across six simulation scenarios and two sample sizes ($N$ = 1,000 and $N$ = 3,000). Each panel corresponds to one scenario. The horizontal axis shows the five machine learning learners. The dashed horizontal line marks the nominal 95% coverage level. Red bars indicate analytical estimates and blue bars indicate bootstrap estimates; lighter shades correspond to $N$ = 1,000 and darker shades to $N$ = 3,000. Note that coverage probability reflects both bias and variance of the estimator; cases where estimates are tightly distributed around an incorrect value (as seen in Figure 2) can still exhibit poor coverage.

## 4.2 Scenario-specific discussion

### 4.2.1 Linear settings – Scenario 1

| *Metric* | | *OLS* | | *LASSO* | | *Random Forest* | | *LightGBM* | | *Neural Networks* | |
|---|---|---|---|---|---|---|---|---|---|---|---|
| | | *N=1000* | *N=3000* | *N=1000* | *N=3000* | *N=1000* | *N=3000* | *N=1000* | *N=3000* | *N=1000* | *N=3000* |
| Bias | | | | | | | | | | | |
| | Analytical | 0.001 | 0.001 | 0.001 | 0.001 | -0.043 | -0.035 | -0.054 | -0.036 | -0.022 | -0.006 |
| Standard Error | | | | | | | | | | | |
| | Analytical | 0.032 | 0.018 | 0.032 | 0.018 | 0.036 | 0.020 | 0.034 | 0.019 | 0.032 | 0.018 |
| | Empirical | 0.034 | 0.018 | 0.034 | 0.018 | 0.033 | 0.018 | 0.036 | 0.019 | 0.033 | 0.018 |
| CI width | | | | | | | | | | | |
| | Analytical | 0.124 | 0.072 | 0.124 | 0.072 | 0.139 | 0.078 | 0.133 | 0.075 | 0.126 | 0.072 |
| | Bootstrap | 0.125 | 0.072 | 0.126 | 0.072 | 0.153 | 0.085 | 0.183 | 0.083 | 0.136 | 0.074 |
| Coverage | | | | | | | | | | | |
| | Analytical | 93.0 | 95.4 | 93.0 | 95.6 | 79.2 | 59.4 | 61.6 | 53.8 | 89.0 | 94.4 |
| | Bootstrap | 94.0 | 96.0 | 94.0 | 95.6 | 79.4 | 52.0 | 71.8 | 34.6 | 80.0 | 92.2 |

**Table 2: Combined Algorithm Performance (Scenario 1, $\boldsymbol{\theta}$ = 0.6)**: Bias is defined as the mean analytical difference between the estimated and true treatment effect ($\theta$ = 0.6) across 500 replications. Coverage refers to empirical coverage probability (%) of the 95% confidence interval. Standard Error rows report the mean analytical SE (derived from DML asymptotic theory) and the empirical SE (standard deviation of estimates across replications). CI width is the mean width of the 95% confidence interval.

In the linear data-generating process (Table 2), all simpler methods have acceptable performance as expected, with the OLS and LASSO performing the best estimating true parameters (having bias of 0.001). On the other hand, Random Forest, LightGBM, and Neural Networks show negative biases that are much larger than OLS and LASSO. In sample size of $N$ =1000, the coverage probability of bootstrap estimates is higher in all cases but Neural Networks. For sample size $N =$ 3000, we observe Random Forest, LightGBM, and Neural Networks having lower coverage for bootstrap. Additionally, OLS and LASSO yield almost identical results, since this simple data generation setting does not contain high-dimensional covariates. Overall, under such a simple data structure, OLS and/or LASSO do better than the machine learning models, in the sense of a lower bias and overall similar mean coverage probability (with coverage probabilities close to 95%). On the other hand, Random Forest and LightGBM undercover substantially (as low as 53.8% at N = 3,000) despite the simplicity of the design.

### 4.2.2 Nonlinear settings – Scenarios 2a and 2b

| *Metric* | | *OLS* | | *LASSO* | | *Random Forest* | | *LightGBM* | | *Neural Networks* | |
|---|---|---|---|---|---|---|---|---|---|---|---|
| | | *N=1000* | *N=3000* | *N=1000* | *N=3000* | *N=1000* | *N=3000* | *N=1000* | *N=3000* | *N=1000* | *N=3000* |
| Bias | | | | | | | | | | | |
| | Analytical | 0.214 | 0.213 | 0.207 | 0.213 | 0.019 | 0.004 | -0.019 | -0.023 | -0.021 | -0.018 |
| Standard Error | | | | | | | | | | | |
| | Analytical | 0.053 | 0.031 | 0.054 | 0.031 | 0.039 | 0.021 | 0.038 | 0.021 | 0.053 | 0.030 |
| | Empirical | 0.053 | 0.032 | 0.055 | 0.032 | 0.034 | 0.018 | 0.037 | 0.018 | 0.044 | 0.033 |
| CI width | | | | | | | | | | | |
| | Analytical | 0.209 | 0.121 | 0.211 | 0.121 | 0.153 | 0.083 | 0.147 | 0.080 | 0.206 | 0.119 |
| | Bootstrap | 0.211 | 0.121 | 0.211 | 0.122 | 0.169 | 0.092 | 0.199 | 0.089 | 0.292 | 0.260 |
| Coverage | | | | | | | | | | | |
| | Analytical | 2.6 | 0.0 | 3.0 | 0.0 | 95.8 | 97.6 | 93.8 | 81.4 | 96.0 | 88.6 |
| | Bootstrap | 2.4 | 0.0 | 2.4 | 0.0 | 97.4 | 98.4 | 98.4 | 71.6 | 100.0 | 100.0 |

**Table 3: Combined Algorithm Performance (Scenario 2a, $\boldsymbol{\theta}$ = 0.6):** Bias is defined as the mean analytical difference between the estimated and true treatment effect ($\theta$ = 0.6) across 500 replications. Coverage refers to empirical coverage probability (%) of the 95% confidence interval. Standard Error rows report the mean analytical SE (derived from DML asymptotic theory) and the empirical SE (standard deviation of estimates across replications). CI width is the mean width of the 95% confidence interval.

For nonlinear settings, the overall performance differs notably from scenario 1. In Scenario 2a ($\theta = 0.6$, Nonlinear, Table 3), OLS and LASSO produce large biases due to the nonlinear terms. In fact, OLS and LASSO have the largest bias (above 0.2) compared with all other methods. On the other hand, Random Forest, LightGBM, and Neural Networks show overall lower biases. Increasing the sample size shrinks the standard deviations and does not have a definitive impact on coverage probability. For example, for LightGBM, increasing the sample size leads to a decrease in coverage probability, from 93.8% to 81.4% for analytical inference and from 98.4% to 71.6% for bootstrap inference; on the other hand, Random Forest benefits from a higher sample size: from 95.8% to 97.6% for analytical inference and from 97.4% to 98.4% for bootstrap inference. Thus, when there is bias, increasing the sample size may reduce variance without eliminating bias, resulting in confidence intervals that are more tightly centered around an incorrect value and potentially harming coverage performance.

| *Metric* | | *OLS* | | *LASSO* | | *Random Forest* | | *LightGBM* | | *Neural Networks* | |
|---|---|---|---|---|---|---|---|---|---|---|---|
| | | *N=1000* | *N=3000* | *N=1000* | *N=3000* | *N=1000* | *N=3000* | *N=1000* | *N=3000* | *N=1000* | *N=3000* |
| Bias | | | | | | | | | | | |
| | Analytical | 0.214 | 0.213 | 0.216 | 0.214 | 0.031 | 0.022 | -0.173 | -0.175 | -0.358 | -0.258 |
| Standard Error | | | | | | | | | | | |
| | Analytical | 0.053 | 0.031 | 0.053 | 0.031 | 0.043 | 0.024 | 0.063 | 0.035 | 0.129 | 0.071 |
| | Empirical | 0.053 | 0.032 | 0.053 | 0.032 | 0.049 | 0.033 | 0.050 | 0.025 | 0.118 | 0.097 |
| CI width | | | | | | | | | | | |
| | Analytical | 0.209 | 0.121 | 0.209 | 0.121 | 0.168 | 0.093 | 0.245 | 0.136 | 0.505 | 0.276 |
| | Bootstrap | 0.211 | 0.121 | 0.211 | 0.122 | 0.233 | 0.157 | 0.404 | 0.191 | 0.937 | 0.841 |
| Coverage | | | | | | | | | | | |
| | Analytical | 2.6 | 0.0 | 2.0 | 0.0 | 87.0 | 74.8 | 17.2 | 0.0 | 18.8 | 8.8 |
| | Bootstrap | 2.6 | 0.0 | 2.2 | 0.0 | 98.6 | 97.8 | 53.0 | 0.0 | 2.6 | 91.4 |

**Table 4: Combined Algorithm Performance (Scenario 2b, $\theta$ = 5):** Bias is defined as the mean analytical difference between the estimated and true treatment effect ($\theta$ = 5) across 500 replications. Coverage refers to empirical coverage probability (%) of the 95% confidence interval. Standard Error rows report the mean analytical SE (derived from DML asymptotic theory) and the empirical SE (standard deviation of estimates across replications). CI width is the mean width of the 95% confidence interval.

Table 4 shows model performance under scenario 2b, where the signal strength increases from 0.6 to 5. We are observing little change in OLS and LASSO, but there are interesting changes in the complex models. First, Random Forest captures the nonlinear relationship effectively, while the estimates of LightGBM and Neural Networks are much weaker compared than in weak signal case in scenario 2a. This occurs because as bias increase, coverage probability decreases accordingly. Note the substantial increase in estimation bias from LightGBM and Neural Networks (from around 0.02 in Table 3 to more than 0.16 in Table 4) are very sensitive to signal strength. Also, we do not observe a systematic pattern between analytical and bootstrap coverages. This occurs for two reasons: The bootstrap CI width can be much wider compared with analytical CI width, and the analytical bias can increase or decrease as we have larger sample sizes.

Overall, comparing between Scenarios 2a and 2b, when there are nonlinear terms and continuous treatment, Random Forest is the least sensitive to signal strength or sample size, and it tends to provide a relatively robust estimate in all cases.

### 4.2.3 Nonlinear and binary settings – Scenarios 3a and 3b

| *Metric* | | *OLS* | | *LASSO* | | *Random Forest* | | *LightGBM* | | *Neural Networks* | |
|---|---|---|---|---|---|---|---|---|---|---|---|
| | | *N=1000* | *N=3000* | *N=1000* | *N=3000* | *N=1000* | *N=3000* | *N=1000* | *N=3000* | *N=1000* | *N=3000* |
| Bias | | | | | | | | | | | |
| | Analytical | 0.192 | 0.227 | 0.211 | 0.245 | 0.095 | 0.062 | -0.071 | -0.017 | 0.027 | -0.034 |
| Standard Error | | | | | | | | | | | |
| | Analytical | 0.281 | 0.162 | 0.281 | 0.162 | 0.190 | 0.096 | 0.155 | 0.080 | 0.210 | 0.070 |
| | Empirical | 0.288 | 0.158 | 0.288 | 0.158 | 0.154 | 0.080 | 0.144 | 0.077 | 0.166 | 0.066 |
| CI width | | | | | | | | | | | |
| | Analytical | 1.102 | 0.635 | 1.101 | 0.634 | 0.744 | 0.376 | 0.608 | 0.312 | 0.824 | 0.275 |
| | Bootstrap | 1.110 | 0.635 | 1.105 | 0.633 | 0.695 | 0.346 | 0.896 | 0.305 | 0.852 | 0.290 |
| Coverage | | | | | | | | | | | |
| | Analytical | 87.2 | 71.4 | 86 | 67 | 96.6 | 92.8 | 95 | 96.4 | 99 | 93.8 |
| | Bootstrap | 87.8 | 72.2 | 86.8 | 66.2 | 97.8 | 97.8 | 99.2 | 95.8 | 99.8 | 70.2 |

**Table 5: Combined Algorithm Performance (Scenario 3a, $\boldsymbol{\theta}$ = 0.6):** Bias is defined as the mean analytical difference between the estimated and true treatment effect ($\theta$ = 0.6) across 500 replications. Coverage refers to empirical coverage probability (%) of the 95% confidence interval. Standard Error rows report the mean analytical SE (derived from DML asymptotic theory) and the empirical SE (standard deviation of estimates across replications). CI width is the mean width of the 95% confidence interval.

Table 5 above shows output of scenario 3a ($\theta$ = 0.6, Binary). In this setting, the estimates from each model are acceptable. However, given that our coverage probability calculation is based on a 95% nominal range, results from LightGBM and Neural Networks are not informative, because their coverage reaches to 99.8%. This suggests an overly wide confidence interval and flawed estimation. On the other hand, Random Forest is slightly better, with its analytical coverage at around the nominal value (96.6% and 92.8%), and bootstrap coverage does not go extremely close to 100%.

| *Metric* | | *OLS* | | *LASSO* | | *Random Forest* | | *LightGBM* | | *Neural Networks* | |
|---|---|---|---|---|---|---|---|---|---|---|---|
| | | *N=1000* | *N=3000* | *N=1000* | *N=3000* | *N=1000* | *N=3000* | *N=1000* | *N=3000* | *N=1000* | *N=3000* |
| Bias | | | | | | | | | | | |
| | Analytical | 0.169 | 0.208 | 0.193 | 0.225 | 0.037 | 0.008 | -0.304 | -0.127 | -0.480 | -0.209 |
| Standard Error | | | | | | | | | | | |
| | Analytical | 0.282 | 0.162 | 0.281 | 0.162 | 0.193 | 0.098 | 0.168 | 0.086 | 0.230 | 0.082 |
| | Empirical | 0.289 | 0.158 | 0.289 | 0.158 | 0.158 | 0.085 | 0.154 | 0.081 | 0.172 | 0.077 |
| CI width | | | | | | | | | | | |
| | Analytical | 1.105 | 0.637 | 1.102 | 0.634 | 0.756 | 0.385 | 0.657 | 0.335 | 0.902 | 0.322 |
| | Bootstrap | 1.108 | 0.635 | 1.106 | 0.634 | 0.728 | 0.370 | 0.901 | 0.355 | 0.984 | 0.467 |
| Coverage | | | | | | | | | | | |
| | Analytical | 89.8 | 75.6 | 86.8 | 70.8 | 97.8 | 98.2 | 55.6 | 66.8 | 43.6 | 27.6 |
| | Bootstrap | 89.6 | 75.2 | 87.6 | 72 | 67.6 | 3 | 94.2 | 6.6 | 16 | 0 |

**Table 6: Combined Algorithm Performance (Scenario 3b, $\theta$ = 5):** Bias is defined as the mean analytical difference between the estimated and true treatment effect ($\theta$ = 5) across 500 replications. Coverage refers to empirical coverage probability (%) of the 95% confidence interval. Standard Error rows report the mean analytical SE (derived from DML asymptotic theory) and the empirical SE (standard deviation of estimates across replications). CI width is the mean width of the 95% confidence interval.

As the signal strength increases in Scenario 3b (Table 6: $\theta$ = 5, Nonlinear Binary), OLS and LASSO are now slightly different, but only in the sense of bias. The standard errors and confidence interval widths are again almost identical as previously seen in scenarios 1, 2a, 2b, and 3a. Also, even though they both show a relatively stable coverage probability in scenarios 3a and 3b, their confidence intervals are the widest among the algorithms we examined: with width above 1 for sample $N = 1000$, and width above 0.6 for sample $N = 3000$. This shows a relatively weak estimation precision, which we believe is because they are incapable of handling the nonlinear complex structures. Other learners are sensitive to signal strength, particularly in their estimation bias (e.g. Random Forest sample $N = 1000$: 0.095 to 0.037 from Table 5 to Table 6). We also observe that generally the estimation standard deviation and confidence interval widths do not show meaningful changes as we look across Tables 5 and 6.

### 4.2.4 Quasi-high dimensional settings – Scenario 4

| *Metric* | | *OLS* | | *LASSO* | | *Random Forest* | | *LightGBM* | | *Neural Networks* | |
|---|---|---|---|---|---|---|---|---|---|---|---|
| | | *N=1000* | *N=3000* | *N=1000* | *N=3000* | *N=1000* | *N=3000* | *N=1000* | *N=3000* | *N=1000* | *N=3000* |
| Bias | | | | | | | | | | | |
| | Analytical | 0.915 | 0.906 | 0.949 | 0.922 | 0.608 | 0.524 | -0.061 | -0.457 | -0.886 | -2.127 |
| Standard Error | | | | | | | | | | | |
| | Analytical | 0.137 | 0.082 | 0.138 | 0.082 | 0.170 | 0.106 | 0.262 | 0.159 | 0.082 | 0.186 |
| | Empirical | 0.148 | 0.078 | 0.149 | 0.079 | 0.122 | 0.069 | 0.240 | 0.142 | 0.066 | 0.106 |
| CI width | | | | | | | | | | | |
| | Analytical | 0.536 | 0.320 | 0.541 | 0.321 | 0.668 | 0.416 | 1.028 | 0.625 | 0.320 | 0.728 |
| | Bootstrap | 0.564 | 0.325 | 0.565 | 0.326 | 0.666 | 0.419 | 1.286 | 0.781 | 0.971 | 0.658 |
| Coverage | | | | | | | | | | | |
| | Analytical | 0 | 0 | 0 | 0 | 0 | 0 | 92.3 | 19.3 | 0 | 0 |
| | Bootstrap | 0 | 0 | 0 | 0 | 0 | 0 | 94.3 | 75.7 | 0 | 0 |

**Table 7: Combined Algorithm Performance (Scenario 4, $\theta$ = 5).** See Table 2 for metric definitions. Based on 300 replications due to computational burden. OLS and LASSO show 0% coverage due to severe bias (~0.91–0.95); LightGBM is the only learner achieving near-nominal coverage at N = 1,000.

Scenario 4 introduces 50 weakly correlated noise predictors to approximate a quasi–high-dimensional setting, where we intend to highlight differences between OLS and LASSO. Based on Table 7, OLS and LASSO still perform similarly, and their coverages drops to 0. Additionally, Random Forest, which we considered the best method from previous scenarios, also has a coverage of 0 for both sample sizes. The only exception in all cases is LightGBM, with a coverage of above 90% for the case $N = 1000$. The case of LightGBM further consolidates our insights: bias could increase as we increase sample size (here analytical bias increases from -0.06 to -0.46). Additionally, Neural Networks fails even more dramatically, as Table 7 shows neural networks bias going from −0.886 to −2.127 with 0% coverage throughout, the worst result in the paper.

# 5. Real data analysis

## 5.1 Data source and preparation

In this section, we apply the DML PLR framework to a real-world county-level dataset constructed by integrating three publicly available U.S. data sources: the Population Level Analysis and Community Estimates (PLACES) data from the Centers for Disease Control and Prevention (CDC) as outcome[23], the Rural–Urban Continuum Codes (RUCC) score from U.S. Department of Agriculture (USDA) for treatment[24], and the demographic survey data from U.S. Census Bureau for covariates[25]. All datasets were collected at the county level to form a unified analytic sample.

The primary outcome is crude county-level adult obesity prevalence from the CDC at year 2021, defined as the percentage of adults with a body mass index (BMI) over 30 kg/m². Our treatment is the 2013 Rural–Urban Continuum Codes (RUCC) score. The RUCC is an ordinal scale from 1 to 9 describing the county-level urbanization score, with lower value representing more urbanized areas and higher values indicating more rural counties. We consider the RUCC in two ways. 1. Using its values of 1-9 as continuous treatment; 2. Dichotomizing into two levels following the USDA definition: counties with RUCC values 1–3 considered as metropolitan areas and those with values 4–9 are non-metropolitan areas. Figure B1 in Appendix B shows a map of the RUCC score (treatment) of each county and the obesity prevalence (outcome).

To adjust for socioeconomic and demographic confounding, we incorporate county-level covariates from the 2013 American Community Survey 5-year estimates[25], including a number of covariates, either directly coming from ACS or derived, shown in Table C1 (in Appendix C). We also include the summary statistics of covariates in Table C2 (in Appendix C). The set of covariates includes measures of income, age structure, sex composition, employment, poverty, marital status, race, education, and health insurance coverage. We use the socioeconomic factors and the rurality, both from 2013, to model the obesity prevalence in 2021. Counties with missing data were excluded from the analysis, and the final dataset had 2,999 complete county-level observations. Our goal is to estimate the causal effect of rurality on obesity prevalence, using 2013 county-level socioeconomic covariates to predict 2021 outcomes under a prospective design.

To incorporate the DML framework, we consider OLS, Random Forest, and Neural Networks to mirror our simulation study. Given the flexibility of Neural Networks, hyperparameters were tuned using the same grid searching method in the simulation settings, over the entire observed dataset. Like our approach in simulations, we did not tune the Random Forest model, as its performance remained robust under default settings.

### 5.2 Results

Table 8 below summarizes the estimated effect of rurality on county-level obesity prevalence, using both continuous and binary treatment.

| *Metrics* | *OLS* | *RANDOM FOREST* | *NEURAL NETWORKS* |
|---|---|---|---|
| **Continuous** | | | |
| Analytical estimate | 0.057 | 0.085 | 0.358 |
| p-value | **0.027** | **0.002** | **<2e-16** |
| Analytical 95% CI | [0.006, 0.107] | [0.031, 0.139] | [0.302, 0.415] |
| Analytical CI width | 0.101 | 0.108 | 0.113 |
| Bootstrap 95% CI | [0.002, 0.106] | [0.026, 0.137] | [0.235, 0.415] |
| Bootstrap CI width | 0.104 | 0.111 | 0.180 |
| | | | |
| **Binary** | | | |
| Analytical estimate | 0.241 | 0.380 | 1.293 |
| p-value | 0.062 | **0.003** | **<4.35e-13** |
| Analytical 95% CI | [-0.013, 0.495] | [0.125, 0.635] | [0.943, 1.643] |
| Analytical CI width | 0.508 | 0.51 | 0.700 |
| Bootstrap 95% CI | [-0.021, 0.488] | [0.115, 0.598] | [0.037, 2.390] |
| Bootstrap CI width | 0.509 | 0.483 | 2.353 |

**Table 8. Estimated Effect of Rurality on County-Level Obesity Prevalence:** Continuous treatment uses the RUCC score (1–9); binary treatment dichotomizes RUCC into metropolitan (1–3) vs. non-metropolitan (4–9). Analytical estimates and CIs are derived from DML asymptotic theory; bootstrap CIs use the percentile method with $B = 100$ resamples. All models use $K = 5$-fold cross-fitting repeated 10 times. Hyperparameters for Neural Networks were tuned using grid search on the full observed dataset. p-values correspond to the two-sided test $H_0: \theta = 0$.

When RUCC is considered continuous treatment, we observe positive and statistically significant analytical estimates across all three methods, indicating an effect between increased rurality and higher obesity prevalence. However, the magnitude of the estimated effect varies substantially by learners. The overall performance of OLS and Random Forest are at a similar level. However, the Neural Networks have markedly larger estimate (0.358) than either OLS (0.057) or Random Forest (0.085), suggesting that Neural Networks may capture more complex structure, or simply tends to overestimate given our small set of covariates. Interestingly, this result is opposite to what we observed in simulation, where the Neural Networks model tends to underestimate the true effect.

When RUCC is considered a binary treatment, we are observing results similar to the continuous case. OLS and Random Forest still have relatively similar interval estimates across analytical and bootstrap methods, with the estimates from Random Forest slightly larger and more significant. On the other hand, Neural Networks have markedly larger estimates.

To interpret the model result, we take the simplest OLS model with binary treatment as an example: The estimated coefficient of 0.241 indicates that, on average, compared to the metropolitan counties (RUCC ≤ 3), non-metropolitan (RUCC ≥ 4) counties have an increasing effect of 0.24 percentage points on adult obesity prevalence, after controlling–for observed socioeconomic and demographic covariates. However, this effect is not statistically significant at the 0.05 level (95% CI: [-0.013, 0.495], p-value 0.062), suggesting marginal evidence of a systematic difference.

In both treatment cases, analytical and bootstrap confidence intervals are nearly identical for the OLS and Random Forest. This indicates that asymptotic variance estimate is not the limiting factor for these two learners. However, bootstrap intervals are substantially wider for Neural Networks. This finding is consistent with our simulation results shown in Section 4.

# 6. Discussion

Our simulation studies show the flexibility of DML to be used in different DGPs, binary or continuous estimation, and adapted with different machine learning methods. In addition, we applied the DML framework to a real-world county-level dataset to assess the association between rurality and obesity prevalence under both continuous and binary treatments. Overall, our results show that even though DML provides a flexible and theoretically solid framework, its practical performance would strongly depend on the true underlying mode, sample sizes, and choice of learners.

Compared to prior evaluations of the DML framework[4], which showcase the robustness of Neural Networks under the DML framework, our results show that Neural Networks can be highly unstable with its large bias and confidence interval width. Our results also connect closely to Bach et al.[9], who studied hyperparameter tuning and practical implementation choices for causal estimation with DML. Bach et al. emphasize that DML performance depends not only on the orthogonalization framework, but also on nuisance learner selection, hyperparameter tuning, and data-splitting strategies. This supports one of the main findings of our study: flexible learners do not automatically guarantee better inference. Therefore, when using flexible machine learning methods, we should pay extra attention to its hyperparameters and model settings, as this could make a huge difference in model performance.

Austin[26] has investigated bootstrap asymptotic variance and bootstrap variance estimation in propensity score weighting methods. They found that bootstrap standard errors are more accurate than asymptotic estimates at smaller samples ($N \leq 1000$), but show no meaningful difference when sample size gets larger, as they tend to get equally decent. In our DML comparison at small sample ($N = 1000$), we are only observing a larger bootstrap coverage for some cases (for all cases with LightGBM, but only one case with Neural Networks). This means sample size is not the only factor in the difference between analytical and bootstrap coverages. Furthermore, increasing sample size sometimes even deteriorates model performance. Another prior work investigated the coverage of variance estimator of doubly robust estimators for average treatment effect. Their work is limited to parametric model settings and thereby focused on whether correctly specifying the model affects coverage. Though the DML PLR model is not strictly a doubly robust estimator, this work incorporates the flexible machine learning models rather than just parametric model, and the coverage results indeed show more variability compared to those of Shook-Sa[27]. Additionally, while this manuscript is being prepared, there are other works that explore the bootstrap properties of the DML estimators. Lin[28] provides a solid theoretical foundation for the bootstrap consistency of DML estimators, and our work can be seen as an applied study for their theory.

There is much more to be done beyond this study. Our study uses the partially linear regression model (PLR), while Chernozhukov also presents the interactive model (IRM) framework. Further comparison can be made between the IRM and PLR. Additionally, the computation burden limits the scope of our simulation comparison, especially with the repeated cross-fitting nested inside hundreds of bootstrap samples. Future research could explore more computationally efficient alternatives or more advanced computation resources to reduce the burden. Finally, while our study considers a set of commonly used machine learning learners, further work could expand this comparison to include more advanced or better-tuned models, such as Deeper Neural Networks or Generalized Additive Models. These extensions would contribute to a more comprehensive understanding of the practical performance of DML methods in both simulated and real-world settings.

# Appendix

## Appendix A. Simulation Grid Tuning results

| *Learner* | *Hyperparameter* | *Candidate Values* |
|---|---|---|
| LightGBM | | |
| | Num_leaves | 15, 31, 63 |
| | Learning_rate | 0.05, 0.10, 0.15 |
| Neural Networks | | |
| | Size | 3, 8, 16 |
| | Decay | 0.1, 0.01, 0.001 |
| | Maxit | 500, 1000 |

**Table A1. Hyperparameter search grid:** Grid tuning was performed only for LightGBM and Neural Networks because these learners showed greater sensitivity to hyperparameter choices in preliminary simulation runs. OLS and LASSO were implemented using their standard linear and regularized regression specifications, while Random Forest was run with default learner settings. For each tuned learner, the final setting was selected based on nuisance prediction performance for the outcome and treatment models within each simulation scenario.

| *Rank* | *num_leaves (g)* | *learning_rate (g)* | *num_leaves (m)* | *learning_rate (m)* | *RMSE_y* | *RMSE_d* | *Score* |
|---|---|---|---|---|---|---|---|
| 1 | 15 | 0.15 | 15 | 0.10 | 1.795 | 0.408 | 2.203 |
| 2 | 15 | 0.15 | 31 | 0.05 | 1.793 | 0.410 | 2.203 |
| 3 | 15 | 0.15 | 31 | 0.15 | 1.780 | 0.424 | 2.204 |
| 4 | 15 | 0.15 | 15 | 0.05 | 1.802 | 0.408 | 2.209 |
| 5 | 15 | 0.15 | 31 | 0.10 | 1.793 | 0.419 | 2.213 |
| 6 | 15 | 0.10 | 31 | 0.05 | 1.804 | 0.411 | 2.215 |
| 7 | 15 | 0.10 | 15 | 0.10 | 1.810 | 0.406 | 2.216 |
| 8 | 15 | 0.15 | 15 | 0.15 | 1.805 | 0.413 | 2.2180 |
| 9 | 15 | 0.15 | 63 | 0.10 | 1.787 | 0.4316 | 2.219 |
| 10 | 15 | 0.15 | 63 | 0.05 | 1.799 | 0.420 | 2.219 |

**Table A2. Top 10 LightGBM Hyperparameter Combinations by Tuning Score (Scenario 3a, θ = 0.6, Nonlinear Binary)***:* g = outcome nuisance model; m = treatment nuisance model. Score = RMSE_y + RMSE_d. Total combinations evaluated: 81 (9 for g × 9 for m). Shown for Scenario 3a as an illustrative example.

| *Rank* | *size (g)* | *decay (g)* | *maxit (g)* | *size (m)* | *decay (m)* | *maxit (m)* | *RMSE_y* | *RMSE_d* | *Score* |
|---|---|---|---|---|---|---|---|---|---|
| 1 | 8 | 0.010 | 1000 | 16 | 0.100 | 1000 | 1.539 | 0.417 | 1.956 |
| 2 | 8 | 0.010 | 500 | 3 | 0.100 | 500 | 1.553 | 0.405 | 1.958 |
| 3 | 8 | 0.010 | 1000 | 8 | 0.010 | 500 | 1.554 | 0.409 | 1.963 |
| 4 | 16 | 0.001 | 500 | 3 | 0.010 | 1000 | 1.561 | 0.406 | 1.967 |
| 5 | 8 | 0.010 | 1000 | 3 | 0.100 | 1000 | 1.565 | 0.403 | 1.968 |
| 6 | 16 | 0.001 | 1000 | 8 | 0.100 | 500 | 1.565 | 0.403 | 1.968 |
| 7 | 8 | 0.010 | 1000 | 3 | 0.001 | 500 | 1.560 | 0.410 | 1.970 |
| 8 | 8 | 0.100 | 1000 | 8 | 0.100 | 500 | 1.567 | 0.403 | 1.971 |
| 9 | 8 | 0.010 | 1000 | 16 | 0.010 | 1000 | 1.539 | 0.433 | 1.972 |
| 10 | 16 | 0.001 | 500 | 3 | 0.001 | 500 | 1.564 | 0.410 | 1.975 |

**Table A3. Top 10 Neural Network Hyperparameter Combinations by Tuning Score (Scenario 3a, $\theta$ = 0.6, Nonlinear Binary)**: g = outcome nuisance model; m = treatment nuisance model. size = number of hidden units; decay = weight decay (L2 regularization); maxit = maximum training iterations. Score = RMSE_y + RMSE_d. Total combinations evaluated: 324 (18 for g × 18 for m). Shown for Scenario 3a as an illustrative example.

# Appendix B. Map of Real Data Analysis: Treatment and Outcome

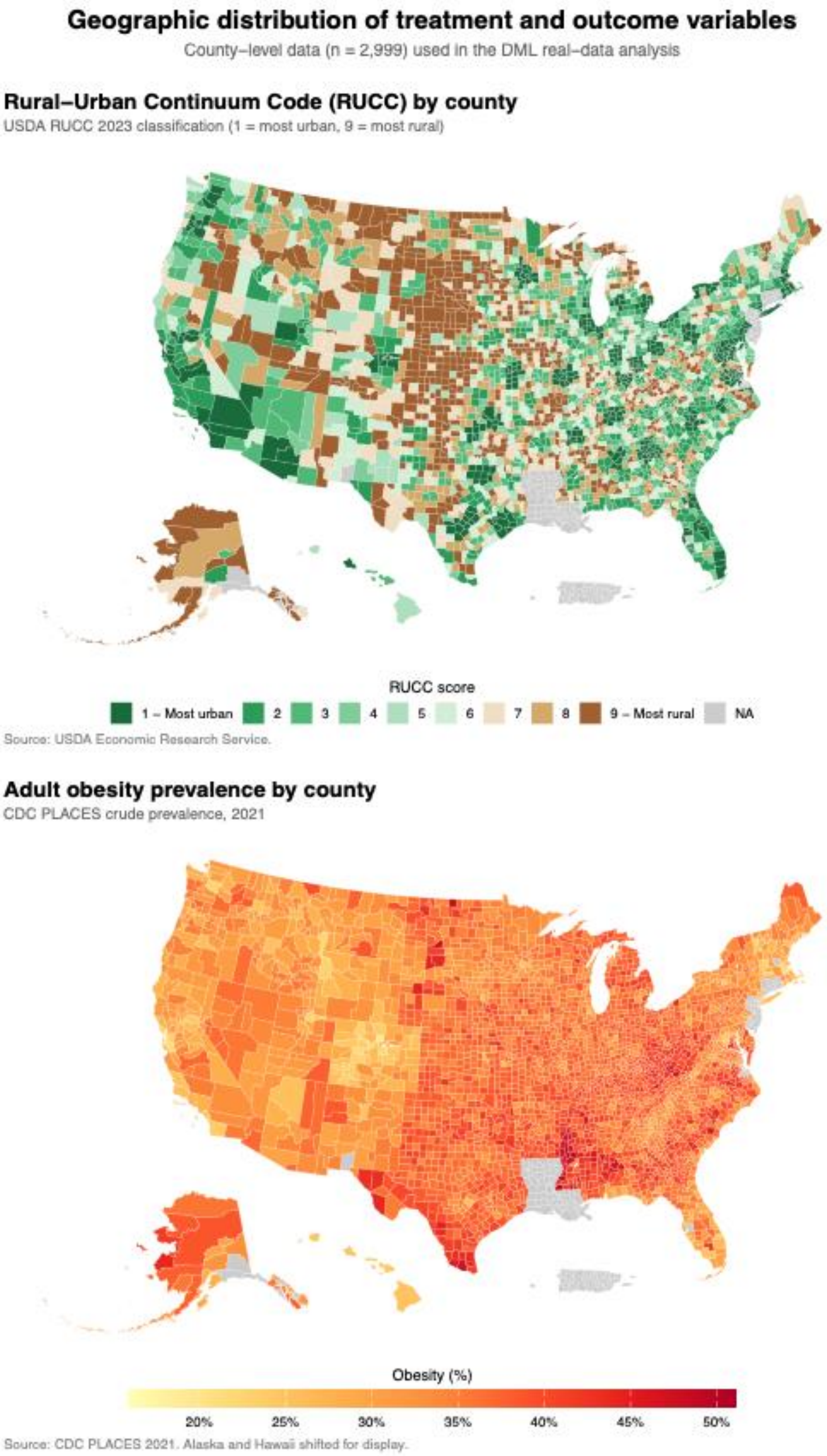

**Figure B1: Geographic distribution of county-level obesity prevalence and rurality (RUCC) across the United States**

## Appendix C. Real Data Analysis Miscellaneous information

| *Covariate* | *Source* | *Type* |
|---|---|---|
| Median household income | Directly extracted from ACS | Direct |
| Median age | Directly extracted from ACS | Direct |
| Female percentage | Female population ÷ total population | Derived |
| Unemployment rate | Unemployed ÷ labor force | Derived |
| Poverty rate | Population below poverty ÷ total population | Derived |
| Marriage rate | (Married males + married females) ÷ population aged 15+ | Derived |
| Black population share | Non-Hispanic Black population (%) | Derived |
| Education rate (Bachelor's or higher) | Sum of Bachelor's-or-higher ÷ population aged 25+ × 100 | Derived |
| Uninsured percentage | Sum of uninsured counts ÷ total population × 100 | Derived |

**Table C1. Construction of County-Level Covariates from ACS 2013:** All covariates are constructed from the 2013 American Community Survey (ACS) 5-year estimates at the county level. Variables labeled "Direct" are used as reported by the ACS, while variables labeled "Derived" are computed using county-level population counts.

| *Variable* | *Overall* | *Metropolitan (RUCC 1–3)* | *Non-metropolitan (RUCC 4–9)* |
|---|---|---|---|
| Number of counties | 2999 | 1074 | 1925 |
| Adult obesity prevalence (%) | 36.1 (33.2, 38.3) | 34.7 (31.8, 37.1) | 36.7 (34.2, 38.8) |
| RUCC score | 6 (3, 7) | 2 (1, 3) | 7 (6, 8) |
| Median household income | 44149 (38343, 50988) | 50332 (43516, 58186) | 41620 (36202, 47169) |
| Median age | 40.7 (37.6, 43.7) | 39.2 (36, 41.5) | 41.7 (38.8, 44.9) |
| Female percentage | 0.50 (0.5, 0.51) | 0.51 (0.5, 0.51) | 0.50 (0.49, 0.51) |
| Unemployment percentage | 0.09 (0.06, 0.11) | 0.09 (0.07, 0.11) | 0.09 (0.06, 0.11) |
| Poverty percentage | 0.16 (0.12, 0.2) | 0.14 (0.11, 0.18) | 0.17 (0.13, 0.22) |
| Black population share | 1.77 (0.49, 8.67) | 4.51 (1.27, 12.99) | 0.97 (0.35, 4.83) |
| Bachelor's degree or higher (%) | 17.51 (13.73, 23.17) | 22.15 (16.77, 29.72) | 16.05 (12.85, 19.82) |
| Uninsured percentage | 3.99 (3.04, 5.14) | 3.80 (2.94, 4.86) | 4.10 (3.09, 5.34) |
| Marriage rate | 0.56 (0.51, 0.59) | 0.55 (0.5, 0.58) | 0.56 (0.52, 0.6) |

**Table C2. Descriptive Summary of County-Level Real-Data by Metropolitan Status :** Values are reported as median (interquartile range), unless otherwise indicated. Metropolitan counties are defined as counties with RUCC scores 1–3, and non-metropolitan counties are defined as counties with RUCC scores 4–9. The analytic sample includes 2,999 counties with complete outcome, treatment, and covariate information.